\documentclass[conference]{IEEEtran}
\IEEEoverridecommandlockouts
\usepackage{cite}
\usepackage{cleveref}
\usepackage{color}
\usepackage{amsmath,amssymb,amsfonts}
\usepackage{algorithmic}
\usepackage{graphicx}
\usepackage{booktabs}
\usepackage{tabularx}
\usepackage{balance}
\usepackage{textcomp}
\usepackage[caption=false]{subfig}
\usepackage[colorinlistoftodos]{todonotes}
\def\BibTeX{{\rm B\kern-.05em{\sc i\kern-.025em b}\kern-.08em
    T\kern-.1667em\lower.7ex\hbox{E}\kern-.125emX}}

\begin{document}
\title{A Preliminary Study on Hyperparameter Configuration for Human Activity Recognition*\\
\thanks{* This work was done in the context of the CONTEXTWA project and was partially funded by the ERDF – European Regional Development Fund through the Operational Programme for Competitiveness and Internationalisation - COMPETE 2020 Programme and by National Funds through the Portuguese funding agency, FCT - Funda\c{c}\~{a}o para a Ci\^{e}ncia e a Tecnologia within project POCI-01-0145-FEDER-016883.}
}
\author{
\IEEEauthorblockN{Kemilly Dearo Garcia, Tiago Carvalho, \\Jo\~{a}o Mendes-Moreira, Jo\~{a}o M.P. Cardoso}
  \IEEEauthorblockA{\small  INESC TEC, Faculty of  Engineering,\\
  University of Porto, Porto, Portugal  \\
    \{kemilly, t.carvalho, jmoreira, jmpc\}@fe.up.pt} \and

  \IEEEauthorblockN{Andr\'{e} C.P.L.F. de Carvalho}%
 \IEEEauthorblockA{\small ICMC, University of S\~{a}o Paulo,\\
 S\~{a}o Carlos, SP, Brazil \\
    andre@icmc.usp.br} }
    
\maketitle

\begin{abstract}  
Human activity recognition (HAR) is a classification task that aims to classify human activities or predict human behavior by means of features extracted from sensors data. Typical HAR systems use wearable sensors and/or handheld and mobile devices with built-in sensing capabilities. Due to the widespread use of smartphones and to the inclusion of various sensors in all contemporary smartphones (e.g., accelerometers and gyroscopes), they are commonly used for extracting and collecting data from sensors and even for implementing HAR systems. When using mobile devices, e.g., smartphones, HAR systems need to deal with several constraints regarding battery, computation and memory. These constraints enforce the need of a system capable of managing its resources and maintain acceptable levels of classification accuracy.
Moreover, several factors can influence activity recognition, such as classification models, sensors availability and size of data window for feature extraction, making stable accuracy a difficult task. In this paper, we present a semi-supervised classifier and a
study regarding the influence of hyperparameter configuration in classification accuracy, depending on the user and the activities performed by each user. This study focuses on sensing data provided by the PAMAP2 dataset. Experimental results show that it is possible to maintain classification accuracy by adjusting hyperparameters, like window size and windows overlap factor, depending on user and activity performed. These experiments motivate the development of a system able to automatically adapt hyperparameter settings for the activity performed by each user.


\end{abstract}


\section{Introduction}

Advanced mobile devices, such as smartphones, are usually integrated with several sensors capable of any-time sensing and data collection. With different types of motions sensors, such as accelerometers, gyroscopes and magnetometers, mobile devices can obtain substantial user-related information by monitoring and tracking movements of their users. Human Activity Recognition (HAR) is a classification task focused on the use of sensing technologies to classify humans activities or to infer human behavior \cite{b1}. Research in this area has been used, for instance, in health care and well-being \cite{b2}, mobile security \cite{b3} and elderly care \cite{b1}.

Although extensive research has been accomplished in the last decade \cite{b1,b2,b4,b5,b6,b7}, most approaches focus on HAR based on supervised learning techniques, assuming the learning data to be always labeled. However, this assumption may not be feasible in real online HAR tasks, when labeled data is rare and the system feedback has to occur at runtime. As an example, in a fall detection system for elderly care, the classification feedback needs to occur as close as possible to the real moment of the user's fall.



To keep a similar accuracy over time, a HAR system needs to adapt itself to the current user. However, due to limitations of most mobile devices, the system needs to manage different resources, such as battery and execution power, to be accurate and functional. Thus, there is a trade-off between amount of processed information and resource management. For example, most work in the literature considers only a fixed window size for all activities and users. As human beings perform activities differently, dissonant input signals are expected for the same activity \cite{b8}.

In this work, we provide an extensive study of the impact of two hyperparameters, window size and overlapping between windows, on activity classification accuracy, for each user, using a leave-one-user-out evaluation approach.
Furthermore, we conducted some experiments with an ODROID-XU+E board in order to evaluate the performance of these hyperparameters in terms of energy consumption and execution time. These experiments allow us to establish a trade-off between classification accuracy and resource consumption, in terms of window size and window overlap.

The rest of the paper is structured as follows.  \Cref{sec:relatedwork} presents related work on HAR and window parameterization. \Cref{sec:metho} describes the methodology applied in this study. \Cref{sec:results} presents and discusses the main results obtained in the experiments. Finally, in \Cref{sec:conclusion}, our final conclusions and future work are summarized.

\section{Related Work}\label{sec:relatedwork}


HAR applications have been extensively researched in recent literature \cite{b4,b5,b6,b7}, specially due to the benefits for human life quality. There are many evidences of the beneficial influence of regular physical activities in reducing problems associated with aging and prevention of various diseases, such as obesity \cite{b2}. Since wearable technologies and smartphones have become more ubiquitous, abundant information about a person's life is available. However, since each person has a unique way of performing an activity \cite{b9}, a HAR system needs to be adapted to the characteristics of a person in order to maintain, or improve, accuracy. In addition, in smartphones devices, it is necessary to manage its limited resources in order to keep the system efficiently working for long periods of time.


Dobbins et al. \cite{b2} propose an approach that uses personal data to better infer lifestyle choices for its users. Considering only labeled data, they evaluate the predictive performance of $10$ supervised HAR classifiers in terms of accuracy and mobile system performance (execution time and energy consumption). They have used an experimental setup based on a fixed window size of $512$ sensor samples with an overlapping window of $0.5$, i.e., $256$ samples are reused from the previous window and only $256$ new samples are used for the current window. They suggest that the sensing data should be processed in the cloud and not in the device. However, personal privacy and Internet connection time are not considered. Furthermore, all data used are labeled, which cannot be guaranteed in a real online mobile system. The datasets used in the experimental section contain complex activities and different user's data, but the results are not compared in terms of accuracy differences per user.


Mannini et al. \cite{b10} proposed the use of an SVM classifier to detect four activities of $33$ different users. The classifier performance was tested for different window sizes, but not the overlap between consecutive windows. It does not compare the classification with different window size in terms of execution time. The results show large variability between users performing the same activity, due to the sensor body location problem. 



Window size has also been discussed by other authors. For example, \cite{b11,b12,b13,b14} compare the predictive performance of classifiers over a set of window sizes. However, most of the studies do not consider the use of overlapping window and the impact of the user on the obtained accuracy.
 
In \cite{b11} an extensive review of the literature in window size and HAR is presented. The accuracy of several classifiers is analyzed for different window sizes, but not regarding users, especially because the experimental setup is not elaborated with a leave-one-user-out experimental methodology. Instead, a cross validation approach is used, which is less affected by user variability than leave-one-participant-out, which would be a more realist approach.

The study conducted in this paper uses the PAMAP2 public dataset \footnote{http://archive.ics.uci.edu/ml/datasets/pamap2+physical+activity+monitoring} \cite{b15}, which includes a vast number of sensors and more complex activities than the experimental data used by many of the studies regarding hyperparameter configurations for HAR. This dataset allowed us to study the impact on HAR accuracy for different window sizes for distinct users and activities.


\section{Activity Recognition Overview}\label{sec:metho}

Sensor data are an important source of information for HAR and current mobile devices usually integrate several types of sensors. As the data from these sensors are user-specific, HAR systems can learn and adapt themselves to their users over time. Since it is not feasible to assume that activity data will be provided with labels, HAR systems need to be able to deal with the situation where none or just a few incoming data are labeled.

\begin{figure}  
  \centering
  \includegraphics[width=0.45\textwidth]
  {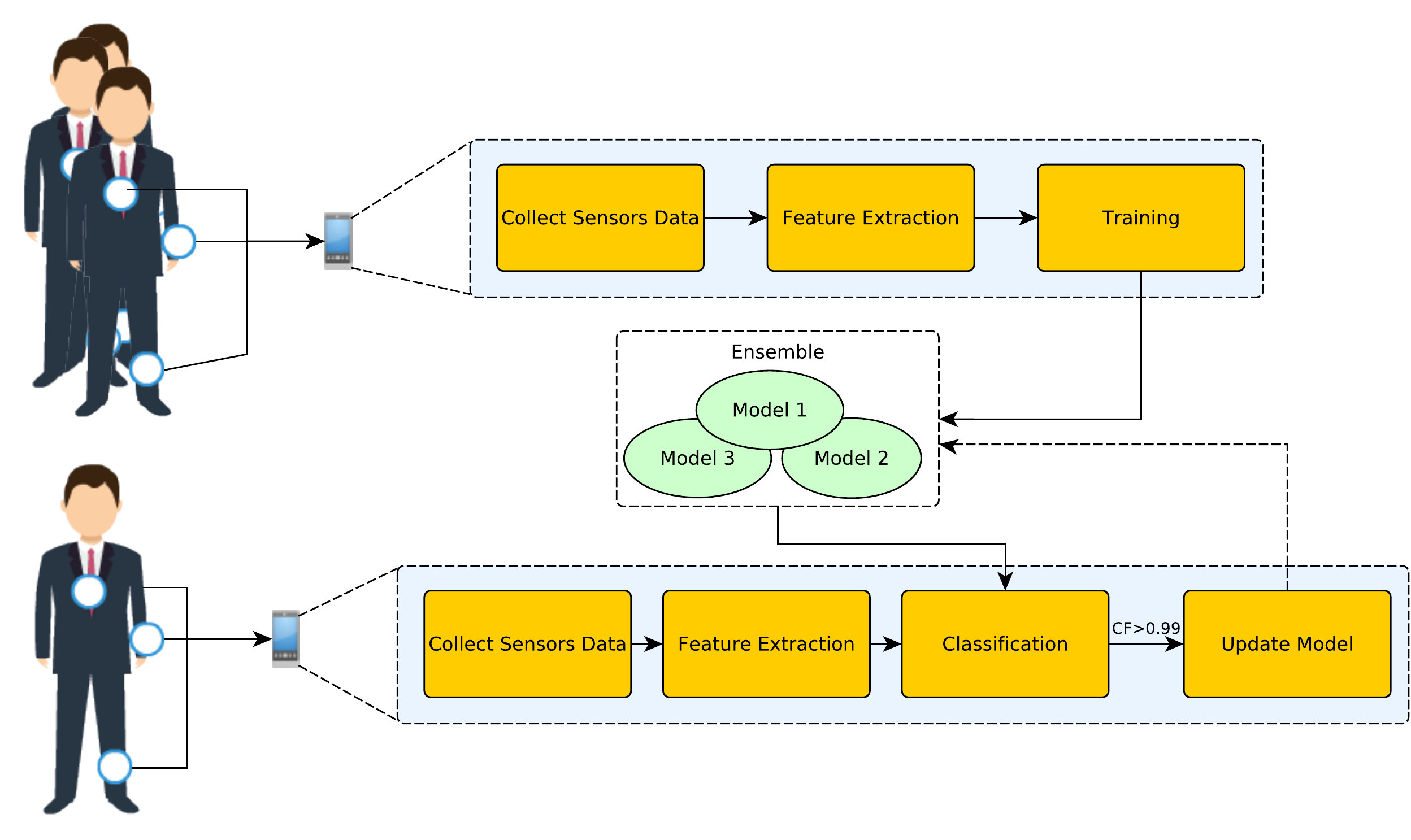}
  \caption{Overview of the semi-supervised system, using an ensemble classifier of three different base classifiers. The models are trained with different users and then specialized for a specific user.}
  \label{Fig:overview}
\end{figure}

In online semi-supervised learning approach, a classification algorithm has access to a dataset where most of the data are unlabeled and can use the labeled data to predict the label for the unlabeled data, as shown in \Cref{Fig:overview}. This process consists of two main phases: (1) an offline phase for training; and (2) an online phase for user-specific classification. 

It starts by training the ensemble classifier with labeled data from several users. A generic version of the classification model is obtained to be used as the basic model. After the training phase and when running the classification for a specific user, the model is updated with user data only if the classification is inferred with a high confidence factor (higher than $99\%$).

The process carried out by this system can be divided into six main steps, the first two steps depict the training phase and the other four steps the online user-specific classification. This work was firstly described in \cite{b15}. The main contribution of this paper is the study on hyperparameter configuration that results from the last task described. The other tasks are herein described for better understanding of the work done.

The raw sensors data is initially collected by a smartphone. Each user has a unique signal to execute the same activity. The more complex the activity performed, more sensors might be required to gather enough information. A typical set of sensors includes a 3D accelerometer, a 3D gyroscope and a 3D magnetometer. However, in this study we use the sensing raw data provided by the PAMAP2 dataset \cite{b15}. 


Then, a sliding window of raw sensor data is processed and converted into a single instance containing the features calculated based on the raw data \cite{b16}. This new instance is used to train (or classify with) an ensemble classifier composed by three classifiers: kNN, VFDT and Naive Bayes. The implementation of the ensemble classifier is a combination of Democratic Co-Learning \cite{b17} and Tri-Training \cite{b18}.

The ensemble training phase is executed in an offline stage, in which the label data is available for training. The label dataset contains raw sensors data from different users performing specifics activities. The first step is the collection of sensors data. 

\Cref{Fig:overview} shows the raw data extracted from different wearable sensors and/or smartphones, placed in different body parts. The raw data samples are stored in a sliding window (a first-in/first-out collection removing the oldest data) which is used in the following step. Since the data from a given classification may be meaningful for the following classification, a between-window overlap factor can be used to reuse samples from the current window in the next window. This means that the new samples collected are reduced by the overlap factor, allowing more classification phases with less collected data.

Since this type of data is usually susceptible to noise, especially accelerometers data \cite{b2}, it is important to process and convert the data into meaningful values. A pre-processing step may also include calibration and filtering of the input signals in order to reduce noise. In a second step, Feature Extraction processes the sliding window to calculate a single instance containing features to be used for the classification. These features (see, e.g., \cite{b16}) include time-domain calculus, specifically mean and standard deviation for each sensor signal and correlation (Pearson correlation) between axes for the 3D sensors. In the online phase, sensors data are collected from a single user and, similarly to the training phase, it can be obtained from wearable sensors and/or smartphones. The second step is also identical to the offline phase.

Each generated instance is classified by the ensemble, which classifies the given instance and provides a confidence factor for that classification. The instances classified with high confidence, more than $99\%$, are used to update the ensemble model. As shown in \Cref{sec:results}, using instances with high confidence factor to update the ensemble model is a powerful mechanism to improve accuracy in HAR systems.

\section{Experimental Results}\label{sec:results}

We conducted several experiments with our approach using the PAMAP2 dataset \cite{b15}. The objective of these experiments is twofold. First, we intend to compare the accuracy of a supervised HAR versus a semi-supervised HAR when using different configurations of two hyperparameters: window size and overlap factor. Then, we intend to study the system behavior with the different hyperparameters configurations in terms of classification accuracy, energy consumption and execution time.

\subsection{The PAMAP2 Dataset}


PAMAP2 \cite{b15} is a public dataset for physical HAR. The data from this dataset was collected from three devices positioned in different body areas: wrist, chest and ankle. Each device contributes with data from three sensors: a 3-axis accelerometer, a 3-axis gyroscope and a 3-axis magnetometer. 

The PAMAP2 dataset contains 18 different activities, which can be divided in basic activities (walking, running, Nordic walking and cycling), posture activities (lying, sitting and standing), everyday activities (ascending and descending stairs), household (ironing and vacuum cleaning) and fitness activities (rope jumping). Also, the users were encouraged to perform optional activities (watching TV, computer work, car driving, folding laundry, house cleaning and playing soccer). 

PAMAP2 contains 1,926,896 samples of raw sensor data from 9 different users. \Cref{tab:activities} shows the number of samples for each activity and user.

\begin{table*}[!ht]
\centering
\caption{Number of samples per activity and user.}
\label{tab:activities}
\begin{tabular}{@{}rrrrrrrrrrrrr@{}}
\toprule
User  & Lying & Sitting & Standing & Walking & Running & Cycling & Nordic Walking & Asc. Stairs & Desc. Stairs & Vac.Clean & Ironing & Rope Jump. \\ \midrule
1 & 27187 & 23480   & 21717    & 22253   & 21265   & 23575   & 20265          & 15890       & 14899        & 22941           & 23573   & 12912      \\
2 & 23430 & 22345   & 25576    & 32533   & 9238    & 25108   & 29739          & 17342       & 15213        & 20683           & 28880   & 13262      \\
3 & 22044 & 28761   & 20533    & 29036   & 0       & 0       & 0              & 10389       & 15275        & 20325           & 27975   & 0          \\
4 & 23047 & 25492   & 24706    & 31932   & 1       & 22699   & 27533          & 16694       & 14285        & 20037           & 24995   & 0          \\
5 & 23699 & 26864   & 22132    & 32033   & 24646   & 24577   & 26271          & 14281       & 12727        & 24445           & 33034   & 7733       \\
6 & 23340 & 23041   & 24356    & 25721   & 22825   & 20486   & 26686          & 13291       & 11272        & 21078           & 37744   & 256        \\
7 & 25611 & 12282   & 25751    & 33720   & 3692    & 22680   & 28725          & 17646       & 11618        & 21552           & 29499   & 0          \\
8 & 24165 & 22923   & 25160    & 31533   & 16532   & 25475   & 28888          & 11683       & 9655         & 24292           & 32990   & 8806       \\
9 & 0     & 0       & 0        & 0       & 0       & 0       & 0              & 0           & 0            & 0               & 0       & 6391       \\ \bottomrule
\end{tabular}
\end{table*}


\subsection{Experimental setup}

Most work in HAR set a signal segmentation in sliding window with fixed size and window overlap. For example, in \cite{b8}, the window size in the experiment was set to 200 samples, which represents four seconds of data collected at a frequency of 50 Hz. In the work presented in this paper we verify experimentally the influence in accuracy of overlap and window size when considering activity and user.

In order to confirm the influence of window size and overlapping in the final classification performance, we use the accuracy metric. The classification accuracy is analyzed by class and user. We train the ensemble classifier with data from eight users and test with data from one isolated user, in a leave-one-user-out approach. For each window, features are processed from raw sensors data resulting in one instance test processed per window.

The experiments were conducted using the PAMAP2 dataset \cite{b15} with the leave-one-user-out validation technique. 
The hyperparameters explored were: the sampling window size (from 100 to 1000 with increments of 100) and the overlapping between windows (from 0.0 to 0.9 with increments of 0.1). 
We conducted four experiments with a HAR system featuring an ensemble classifier consisting of three classifiers: kNN, Naïve Bayes, and Hoeffding Tree (VFDT), as in \cite{b8}. As verified in \cite{b2}, these three classifiers have good classification performance.

The box-plots correspond to the ensemble accuracy over the changes in the window size and overlapping. The heat-map tables are included to facilitate the visualization in terms of how the hyperparameters influence the classification of each activity for each user.

We also intend to analyze the impact of using a semi-supervised approach for activity recognition with a leave-one-user-out implementation. Considering that each user produces different input signals for the same activity, hence influencing the accuracy of the classifier, it is necessary for the classifier to adapt itself to the user.
We show accuracy results by: 
window size, overlapping size, user and activity.


\begin{figure}
  \centering
  \includegraphics[width=0.5\textwidth]{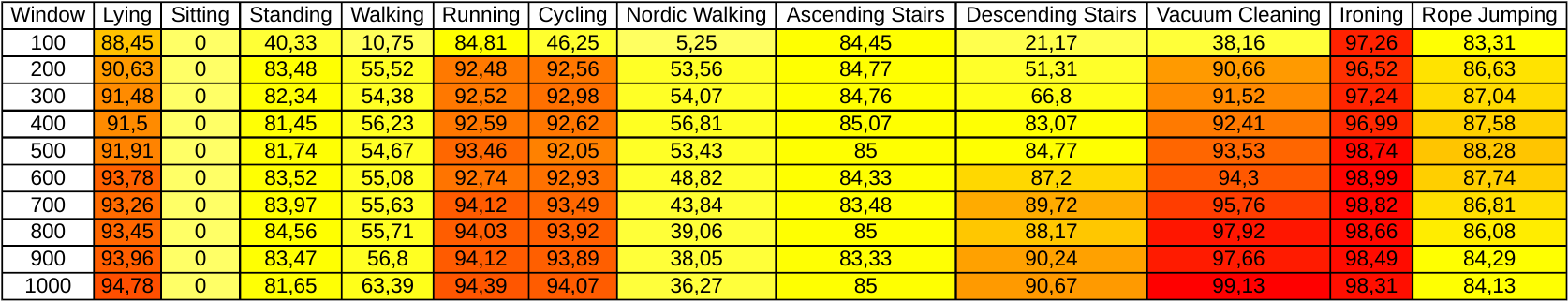}
  \caption{Semi-supervised Ensemble accuracy, for user 1, overlapping 0.8 varying window size.}
  \label{Fig:table user 1 O8}
\end{figure}

\begin{figure}
  \centering
  \includegraphics[width=0.45\textwidth]{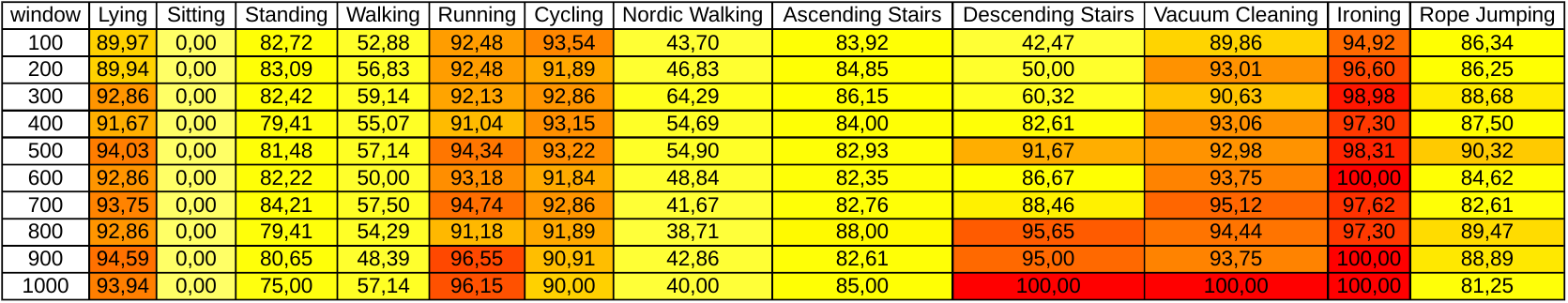}
  \caption{Semi-supervised Ensemble accuracy, for user 1, overlapping of 0.2 varying window size.}
  \label{Fig:table user 1 O2}
\end{figure}

\begin{figure}
  \centering
  \includegraphics[width=0.45\textwidth]{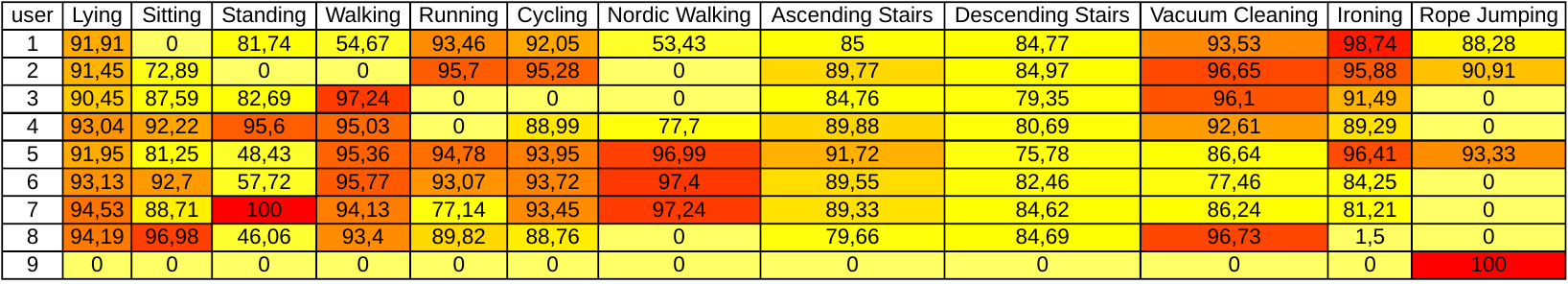}
  \caption{Semi-supervised Ensemble accuracy, varying user, window size 500 and overlapping 0.8.}
  \label{Fig:table user}
\end{figure}



\begin{figure}
 \centering
	\subfloat[Semi-supervised]{
    \includegraphics[width=0.41\textwidth]{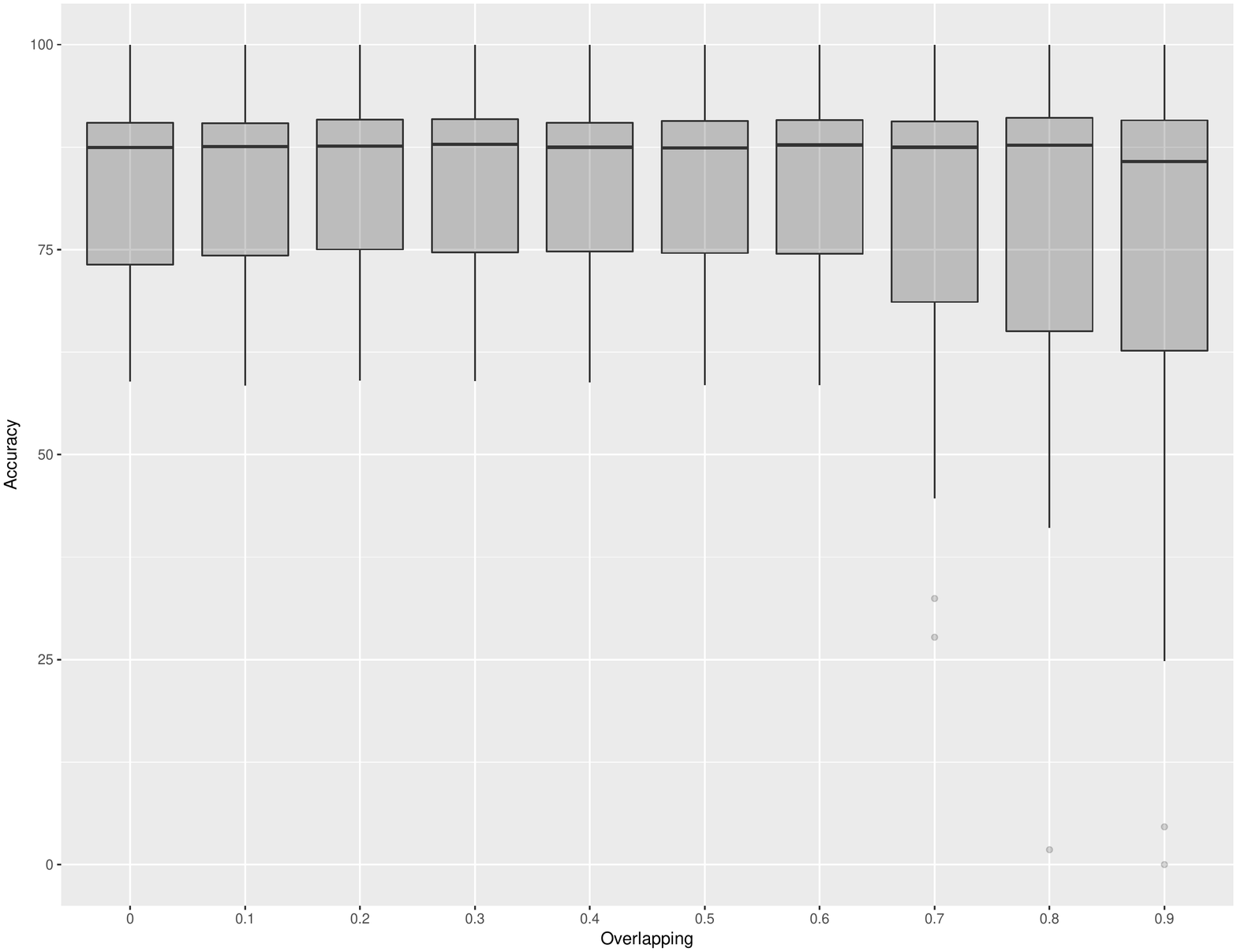}
    }
    \hfill
    \subfloat[Supervised]{
    \includegraphics[width=0.41\textwidth]{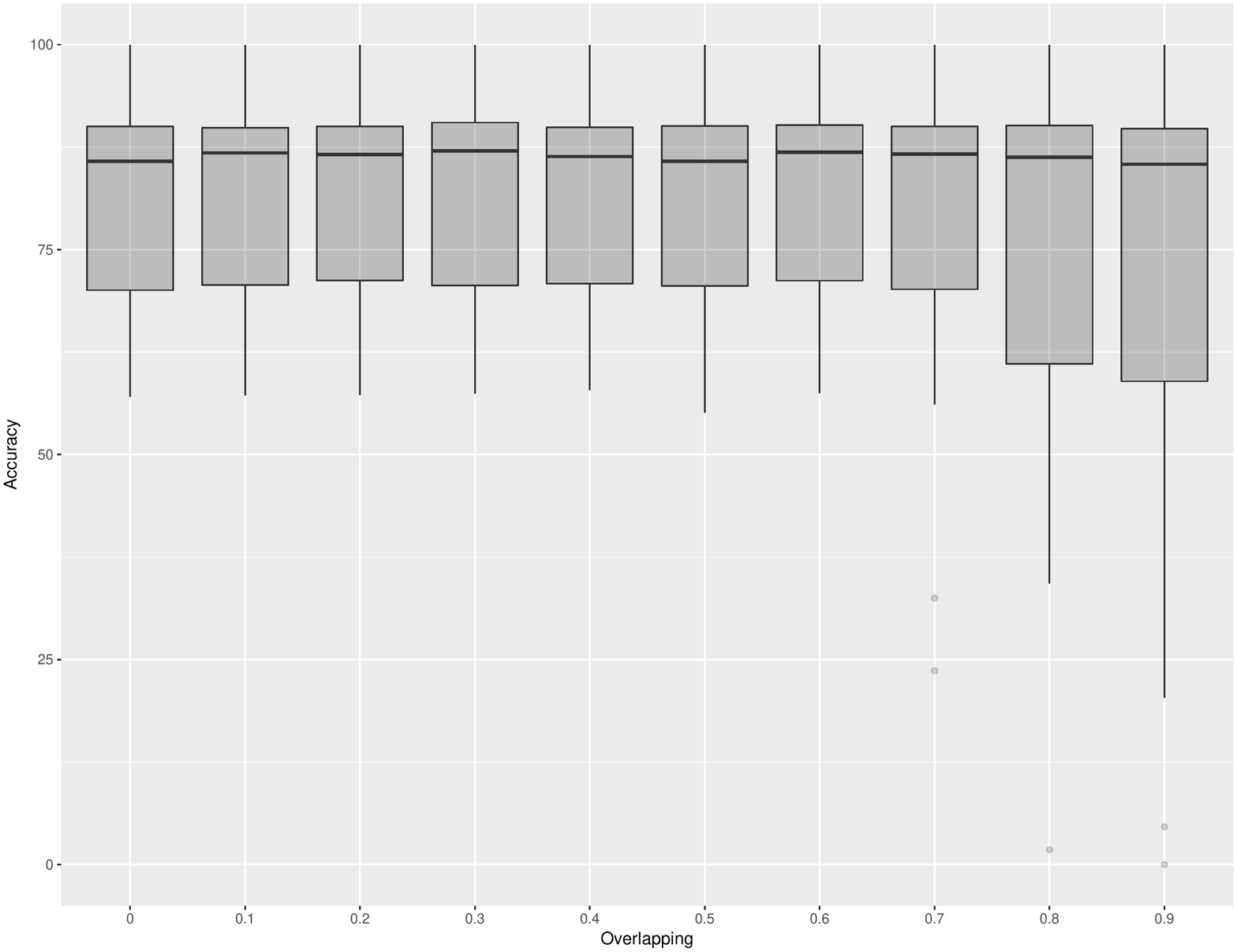}
    }
  \caption{Supervised vs Semi-Supervised Ensemble accuracy: Overlap influence.}
  \label{Fig:overlap}
\end{figure}

\begin{figure}  
  \centering
  \subfloat[Semi-supervised]{
  	\includegraphics[width=0.41\textwidth]{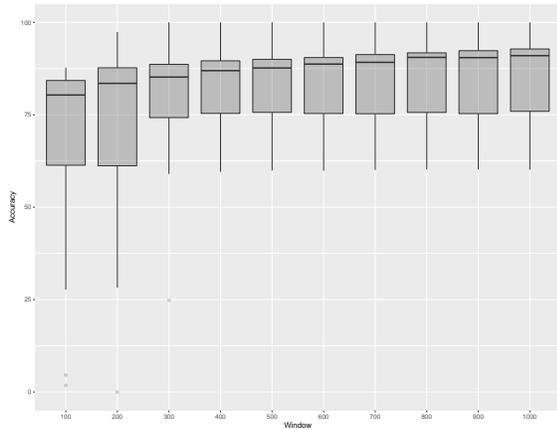}
  }
	\hfill
  \subfloat[supervised]{
  	\includegraphics[width=0.41\textwidth]{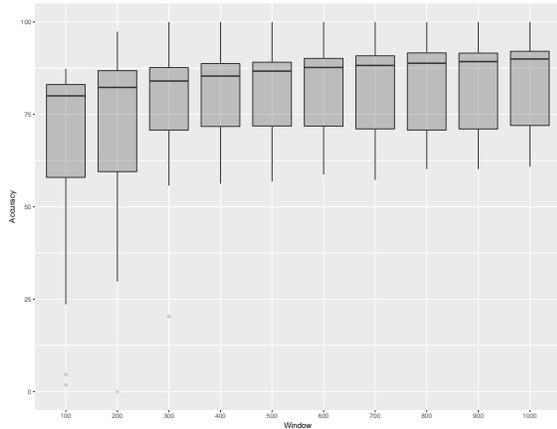}
  }
  \caption{Supervised vs Semi-Supervised Ensemble accuracy: window size influence.}
  \label{Fig:window}
\end{figure}

\begin{figure}
 \centering
	\subfloat[Semi-supervised]{
  \includegraphics[width=0.41\textwidth]{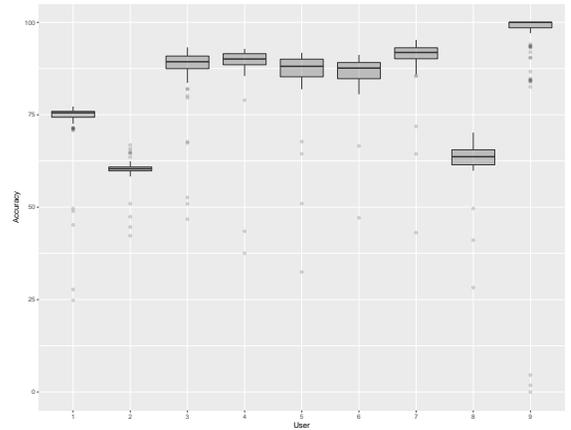}
  }
  \hfill
  \subfloat[Supervised]{
  	\includegraphics[width=0.41\textwidth]{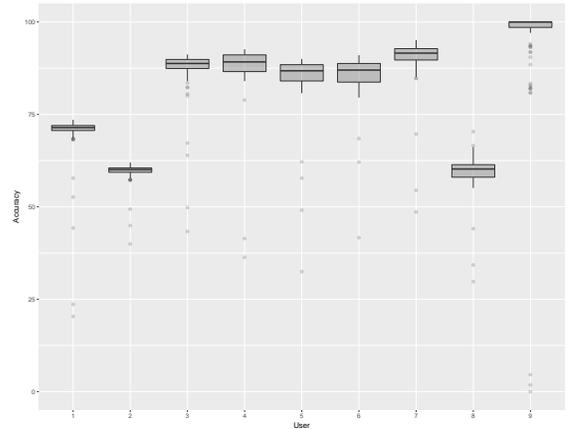}
  }
  \caption{Supervised vs Semi-Supervised Ensemble accuracy: user influence.}
  \label{Fig:user}
\end{figure}




\subsection{HAR Accuracy Results}\label{sec:discuss}


Semi-supervised can improve accuracy, but not significantly. In \Cref{Fig:overlap} the semi-supervised approach reduces variance for most of overlapping window and has average accuracy close to 90\%. Overlapping has more influence on accuracy for values higher than 0.7, but semi-supervised is less susceptible to that influence than the supervised approach. Semi-supervised also improves accuracy for window size hyperparameter, as shown in \Cref{Fig:window}. Windows with small sizes has worse results, especially for sizes equal to 100 and 200. 

In \Cref{Fig:user}, we analyze the impact of accuracy when the hyperparameters of window size and overlapping for each user are changed. For user 5, either in supervised approach or semi-supervised, the variance is higher than users 2 and 1. The semi-supervised ensemble reduces accuracy variance in users 4 and 8. 

User 9 is an interesting case to analyze. For most of the cases, the accuracy is 100\%, however, as is shown in \Cref{tab:activities}, user 9 only has instances for the Rope Jumping activity, which means that this user influences the results to higher values, but in comparison with users 2 and 1 for some cases the individual accuracy can be lower, justifying the analyses by user instead of analyzing all population. 

\Cref{Fig:table user} is a heatmap table in which the highest values are red, the lowest are yellow and the intermediate values are orange. For window size 500 and overlapping of 0.8, the accuracy of each activity has notable differences depending on the user. For example, for user 1 the highest accuracy is for the activities Running, Cycling, Vacuum Cleaning and Ironing, but is not the same for user 7, where activities, such as: Lying, Standing, Walking and Nordic Walking, have better accuracy values. Both users have a significant amount of instances for each activity, as seen in \Cref{tab:activities}.

\Cref{Fig:table user 1 O8} and \Cref{Fig:table user 1 O2} show the impact on accuracy per class over user 1. 
\Cref{Fig:table user 1 O8} has overlapping window of 0.8 and \Cref{Fig:table user 1 O2} has overlapping window of 0.2.
With overlapping of 0.2, activity Cycling is less affected to window size than with overlapping 0.8. However, \Cref{Fig:table user 1 O8} shows higher accuracy results. This is noticeable for the others activities too. Especially for window size 100, accuracy values are higher for most of the activities with overlapping 0.2.

As expected we can see with these results that window size and window overlapping do influence the accuracy. Based on these results and depending on the HAR application, one can decide about the window size and overlapping level that makes possible a certain minimum desired classification accuracy. The exhaustive exploration allows us to also understand the acceptable ranges to explore within a runtime autotuning system, e.g., to keep a minimum accuracy (e.g. 80\%). These ranges can be used, at runtime, to search for the best combination of the hyperparameters that provide the best results, e.g., in terms of execution time or energy consumption. The following subsection shows the impact on execution time and energy consumption of different window sizes and overlapping levels.



\subsection{Time and Energy Consumption}

We conducted some experiments in an ODROID-XU+E \footnote{ODROID-XU+E is a board mainly consisting of an Exynos5 Octa SoC, which includes 2 quad cores ARM CPUs and a PowerVR GPU, and a power measurement circuit to measure CPU, GPU and DRAM power consumption. The Exynos5 Octa SoC has been used in a number of families of smartphones.}
\footnote{https://www.hardkernel.com/} 
system running Android to observe the execution time and energy consumption for processing all the data from the PAMAP2 dataset. The experiments focus on a single user, user 6 in \Cref{tab:activities}, and the execution time and energy required to process the 250,096 raw samples. We also limit the exploration space of window size and overlap factor to a narrowed range of values that keep the classification accuracy under certain limits.

\begin{figure}  
  \centering
  \includegraphics[width=0.40\textwidth]{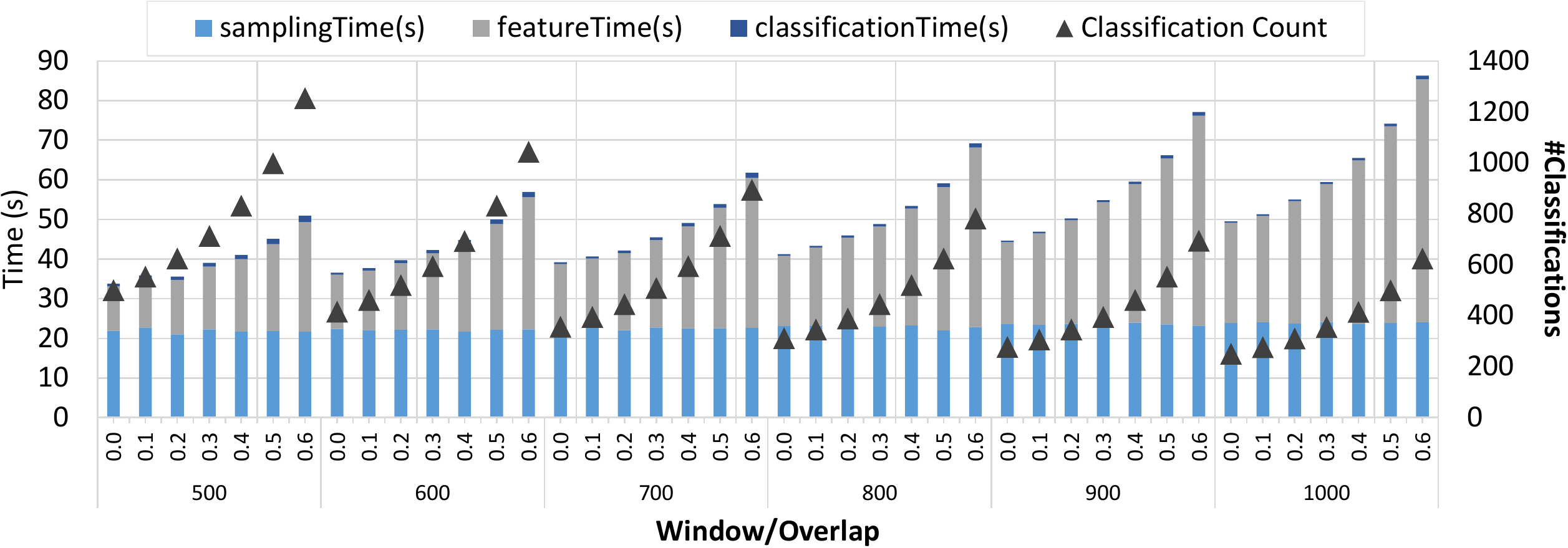}
  \caption{Total execution time required (left axis) to process the PAMAP2 dataset, per window size and overlap factor, divided in three parts: sampling (data extraction), features extraction and classification. The number of classifications per configuration (right axis) is shown as triangle marks.}
  \label{Fig:time}
\end{figure}

\Cref{Fig:time} shows the execution time necessary to process all the data of user 6 and with the time divided into three parts. The first part, \texttt{samplingTime}, represents the time required to access all the data from the user and the instantiation of each data window as an instance. The second part, \texttt{featureTime} is the feature extraction and ``final instance'' instantiation and  depends on the window size and overlap factor used. The last part, \texttt{classificationTime} is the total time required to classify all the instances calculated in the feature extraction phase.

Since the feature extraction depends on the window size, the time to calculate all instances increases as the window size also increases, despite the decreasing number of calculated features. This means that the feature extraction phase is sensitive to the number of raw instances to process. Furthermore, as we increase the overlap factor, due to the increased number of instances that are calculated, the time also increases. The classification time is rather small and slightly increases as the number of calculated features augments. 

\begin{figure}  
  \centering
  \includegraphics[width=0.45\textwidth]{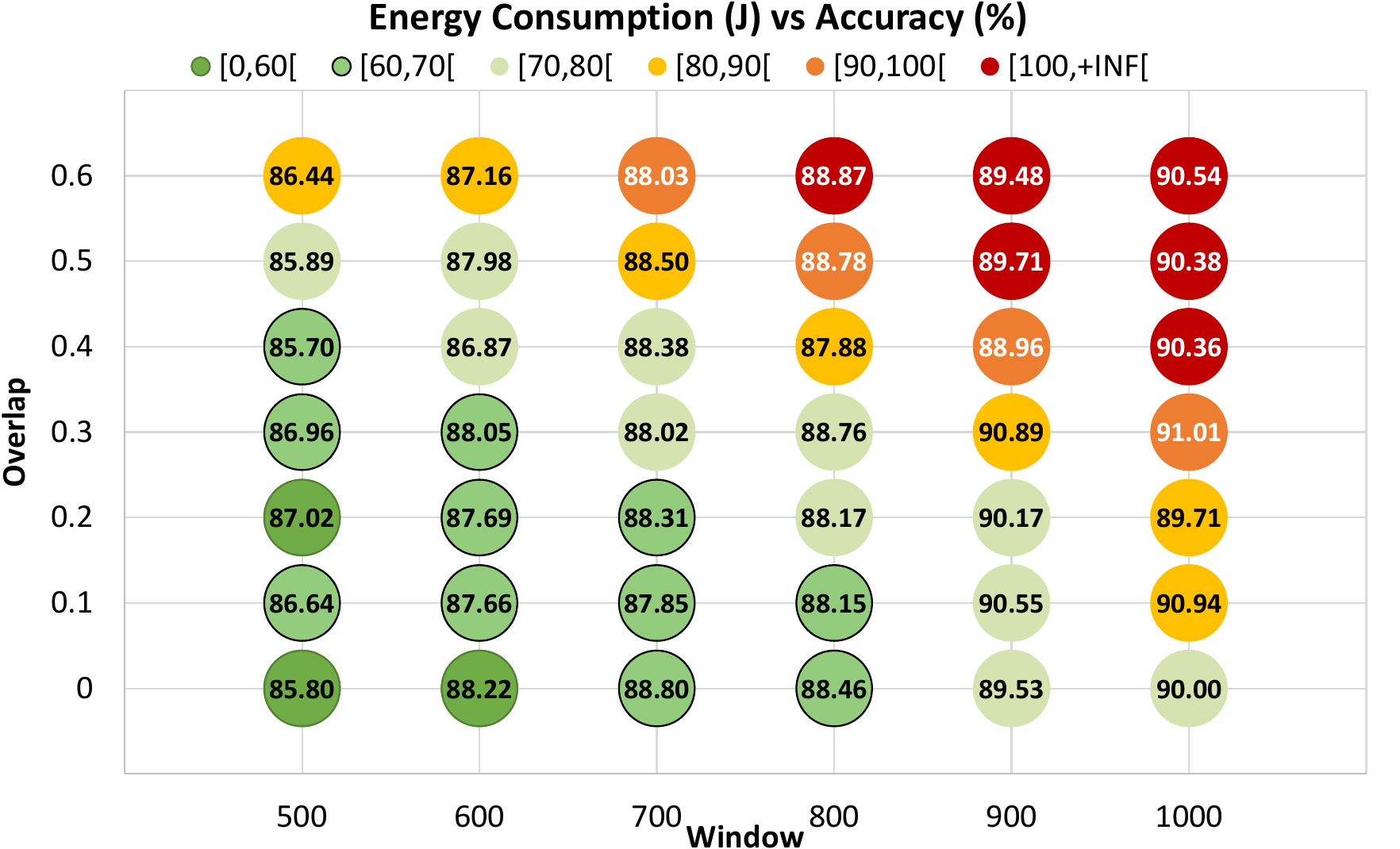}
  \caption{Energy consumed, in Joules, while processing the PAMAP2 dataset, per window size and overlap factor. Green values represent less energy consumed while red values represent higher energy consumed. The values in each configuration represent the accuracy, in percentage, of that configuration.}
  \label{Fig:energy}
\end{figure}

\Cref{Fig:energy} shows a heat map representing the energy (J) spent to process the PAMAP2 raw data for the different window sizes and overlap factors. The colors represent a range of Joules, where the red color depicts more energy consumption and, reversely, green color depicts less energy. The accuracy is also shown in the map over each circle depicting the energy color to compare the energy spent with the classification accuracy.

We can see that smaller windows result in less energy consumption than bigger windows. This may be due to the increased effort to calculate features for larger window sizes. It is also perceivable that increasing the overlap factor also increases the energy spent, essentially due to the increased number of feature calculations and classifications to be carried out.
Relating the energy consumption with the accuracy achieved for a given configuration, it is observable that the best accuracy values reside in more ``heated'' zones, i.e., where energy consumption is higher. Lower window sizes present less accuracy while higher window sizes provide more accuracy. For instance, in configurations without overlapping (i.e., with an overlap factor of 0), the accuracy rises from 85\% for a window of size 500 to 90\% for a window of size 1000.
The overlap shows more fluctuations in terms of accuracy, however with the best factors concentrated between 0.1 and 0.5. This shows that it is not trivial to select a single window size and overlap factor if it is intended to have two possible scenarios, one where accuracy is the most important factor and another one where energy consumptions is the top priority but still with a minimum accuracy  value in mind.

\section{Conclusion}\label{sec:conclusion}

Human activity recognition (HAR) is a classification task for which raw sensors data are crucial. Humans perform activities in unique ways reflected in distinct signals from sensors, which influence accuracy performance. In this paper we presented an analysis of the impact of hyperparameters as window size and overlapping on classification accuracy, execution time and energy consumption. The analysis was focused on a public dataset, which includes raw sensor data from different users.


The results provided confirm the need of adapting the classification model to the current user.
Due to the impact of window size and overlapping, each activity requires a specific configuration of these hyperparameters in order to improve classification accuracy.
Furthermore, the results also motivate the development of a system that is able to adapt the application at runtime when trade-offs between performance accuracy and energy consumption need to be considered. Bearing in mind this, the window size and overlap factor can be used to develop runtime strategies able to adapt these parameters according to the target goals.

Ongoing work focuses on further studies considering other hyperparameter configurations, as the sampling rate and classifier parameters (e.g., the value of k in kNN). As future work we plan to implement a system able to dynamically adjust at runtime the window size and overlapping factor and aware of activities and users. The dynamic adaptation needs to consider an exploration of possible parameter configurations to find the best configurations for each adaptation scenario and thus the experimental results presented in this paper are also part of that exploration phase.

\end{document}